# Système d'aide à l'accès lexical : trouver le mot qu'on a sur le bout de la langue

Lortal G., Grau B. et Zock M.
LIMSI-CNRS – Université Paris XI
BP 133 91403 Orsay
gaelle.lortal@utt.fr, brigitte.grau@limsi.fr, zock@limsi.fr

## Résumé - Abstract

Le Mot sur le Bout de la Langue (Tip Of the Tongue en anglais), phénomène très étudié par les psycholinguistes, nous a amené nombre d'informations concernant l'organisation du lexique mental. Un locuteur en état de TOT reconnaît instantanément le mot recherché présenté dans une liste. Il en connaît le sens, la forme, les liens avec d'autres mots... Nous présentons ici une étude de développement d'outil qui prend en compte ces spécificités, pour assister un locuteur/rédacteur à trouver le mot qu'il a sur le bout de la langue. Elle consiste à recréer le phénomène du TOT, où, dans un contexte de production un mot, connu par le système, est momentanément inaccessible. L'accès au mot se fait progressivement grâce aux informations provenant de bases de données linguistiques. Ces dernières sont essentiellement des relations de type paradigmatique et syntagmatique. Il s'avère qu'un outil, tel que SVETLAN, capable de structurer automatiquement un dictionnaire par domaine, peut être avantageusement combiné à une base de données riche en liens paradigmatiques comme EuroWordNet, augmentant considérablement les chances de trouver le mot auquel on ne peut accéder.

The study of the Tip of the Tongue phenomenon (TOT) provides valuable clues and insights concerning the organisation of the mental lexicon (meaning, number of syllables, relation with other words, etc.). This paper describes a tool based on psycho-linguistic observations concerning the TOT phenomenon. We've built it to enable a speaker/writer to find the word he is looking for, word he may know, but which he is unable to access in time.
We try to simulate the TOT phenomenon by creating a situation where the system knows the target word, yet is unable to access it. In order to find the target word we make use of the paradigmatic and syntagmatic associations stored in the linguistic databases. Our experiment allows the following conclusion: a tool like SVETLAN, capable to structure (automatically) a dictionary by domains can be used sucessfully to help the speaker/writer to find the word he is looking for, if it is combined with a database rich in terms of paradigmatic links like EuroWordNet.

## Keywords – Mots Clés

MBL, accès lexical, relations sémantiques, associations, SVETLAN, EWN
TOT, word access, semantic relations, associations, SVETLAN, EWN

# 1 Introduction

Qui ne s'est pas encore trouvé dans cette situation frustrante à chercher un mot qu'il connaît, sans être en mesure de trouver sa forme pour autant ? Le mot étant connu, il est donc disponible, mais n'arrivant pas à y accéder, il vous reste collé sur le bout de la langue ? Ce phénomène observé depuis l'antiquité (Aristote en parle déjà) a été étudié maintes fois par des psychologues (Brown & Mc Neill, 1966, Aitchinson 1997). Ces travaux révèlent que les personnes se trouvant dans cet état savent non seulement énormément de choses concernant le mot recherché (mot cible) (Brown 1991)— origine, tout ou partie de son sens, nombre de syllabes, lien avec des mots associés, etc.— mais les mots avec lesquels ils le confondent (mots sources) ressemblent étrangement au mot cible : lettres ou initiale, catégorie syntaxique, champ sémantique, etc. L'état de TOT n'est pas l'oubli d'un mot. On ne perd pas l'information en mémoire. Le TOT n'est qu'un exemple d'échec de récupération d'information en mémoire. L'information est disponible sans être accessible à un moment donné.

Nous présenterons ici les premiers résultats d'un travail dont l'objectif est de réaliser un programme tirant bénéfice de cet état de faits pour assister un locuteur (ou rédacteur) à (re)trouver le mot que celui-ci a sur le bout de la langue. Autrement dit, notre objectif est de créer, à terme, un dictionnaire analogue à celui de l'être humain (réseau associatif, dont chaque terme est indexé de manière multiple) [1]. Un tel dictionnaire permettrait donc l'accès à l'information recherchée soit par la forme (lexicale : analyse), soit par le sens (concepts : production), soit par les deux, simultanément, ou l'un après l'autre [2].

Notre point de départ est identique à celui de Zock (2002, 2004) et Zock et Fournier, (2001) qui, pour trouver le mot recherché, s'appuient sur la notion de stockage bi-modal (forme et sens), idée fondée sur l'analyse des erreurs de discours spontané (Fromkin, 1973). A l'instar des êtres humains, l'accès par la forme peut se faire de deux façons, par la graphie ou le son. L'accès par le sens se fonde en grande partie sur la notion d'associations. Le dictionnaire mental étant un hyper-réseau dont les noeuds correspondent aux mots et les liens à des associations, trouver un mot dans ce réseau consiste alors à entrer à n'importe quel point, puis à suivre les liens. Cela suppose néanmoins qu'un tel réseau associationniste soit construit au préalable. Idéalement, on s'appuierait donc sur un inventaire exhaustif de liens entre les mots (liens associatifs), liens qui seraient pondérés par la force associative (fréquence relative des couples de mots). Voilà l'objectif de ce travail : identifier un mot manquant au sein d'un texte (corpus), selon sa place dans la phrase et des points d'entrée dans le réseau (les différents types d'accès cités ci-dessus). Disposant actuellement de deux ressources sémantiques de nature très différente, SVETLAN (Chalendar et Grau, 2002) et EuroWordNet (EWN, Vossen, 1999), nous verrons si leurs points d'entrée sont suffisants pour atteindre notre objectif.

---

[1] Pour être plus précis il ne s'agit pas de créer un dictionnaire de toute pièce, il s'agit plutôt d'enrichir une base de données existante (TLF, WordNet) avec des liens typiques pointant d'un mot aux autres fréquemment associés.

[2] Il arrive qu'on ne trouve pas d'emblée le bon candidat, auquel cas on part d'un mot raisonnablement proche (synonyme, hyperonyme, antonyme), espérant par ce biais de se rapprocher du mot cible.

## 2  Les relations sémantiques et lexicales

### 2.1  Les relations dans le cadre du TOT

L'étude du TOT montre qu'il existe des relations de différents niveaux : phonétique, graphémique, sémantique. Un locuteur en train de chercher un mot fait donc appel à différents types de mémoires, à différents systèmes interconnectés (Le Rouzo et Joubert, 2001). La mémoire verbale est composée de la mémoire sémantique et de la mémoire lexicale ; cette dernière contient des mots.

La mémoire sémantique contient des concepts, le sens des mots et des symboles. La récupération du signifié se ferait alors par un mot ayant un sens proche du mot cible, en suivant des relations paradigmatiques. La mémoire lexicale concerne la forme des mots, leur graphie ou leur prononciation. La récupération du signifiant se ferait alors par des mots contenant des sons semblables, les mêmes lettres (initiales, finales) ou encore le même nombre de syllabes que le mot cible.

A l'intérieur de ces composantes de la mémoire à long-terme, deux types de récupération sont à distinguer : le rappel et la reconnaissance. Dans le cas du rappel, l'information est reproduite de la mémoire. La catégorisation ou la structuration de l'information assiste le rappel (relations paradigmatiques du signifié par exemple ou relations sur le signifiant). La reconnaissance est liée à la présentation d'un stimulus évoquant une expérience passée constituant un amorçage et entraînant l'utilisation de liens associatifs.

### 2.2  Les relations linguistiques

On peut corréler les différents types de récupération lexicale avec les différents types de relations linguistiques ou conceptuelles. Deux grandes catégories de relations sont considérées : les relations paradigmatiques et leur pendant, les relations syntagmatiques.

- Les relations paradigmatiques

Les relations paradigmatiques sont des relations de similarité (synonymie, antonymie) et d'inclusion (hyponymie, méronymie, etc.). Dans « l'oiseau vole dans le ciel bleu », « *oiseau* » entretient un lien d'hyperonymie avec « *aigle* », et les mots « oiseau, vole, ciel, bleu » pourraient bien être remplacés par leurs synonymes. Ces types de relation ont été très étudiés et constituent l'essentiel de WordNet et donc de la version française d'EuroWordNet.

- Les relations syntagmatiques

Les relations syntagmatiques sont les liens qu'une unité linguistique possède avec d'autres unités présentes dans un énoncé (café noir, café du Brésil, café fort). Elles correspondent aux associations, la présence d'un terme évoquant l'autre. Ces relations permettent un amorçage sémantique : l'accès à un mot est facilité s'il a été précédé par un autre mot qui lui est sémantiquement associé.

Ces relations varient avant tout en fonction des expériences personnelles du locuteur. Mais elles varient également en fonction du sociolecte, de l'idiolecte et même en fonction du contexte (analogies, co-occurrences, collocations). Par exemple, les collocations indiquent un lien syntagmatique arbitraire et figé, lien perdurant dans la mémoire. Ce type de lien a été très étudié dans la Théorie Sens-Texte de (Mel'cuk et al.,1995) grâce à la notion de fonction lexicale. Une fonction lexicale (FL) exprime le fait que le choix d'un mot X dépend non seulement de son sens (comme on pourrait l'attendre), mais aussi d'autres mots auxquels il est typiquement associé. Prenons le domaine des sentiments. Dans « colère noire » le mot « noire » n'exprime pas la notion de couleur, mais celle d'intensité forte, tandis qu'en cas de « peur » les mots convenables seraient « bleue » , « grande» ou « très», puisqu'on dit « peur bleue », « grande peur», « très peur», et non « forte peur ».

Cette brève incursion dans le domaine du TOT révèle que le problème en question est soluble en ayant recours à des liens formels (phonologique et graphique) et sémantiques. Comme indiqué, nous disposons de deux types de liens, des liens explicites pour les liens paradigmatiques dans le cas d'EWN (Vossen, 1999) et implicites dans les domaines de SVETLAN pour les liens syntagmatiques. Pour des raisons d'espace nous ne présenterons ici que ce dernier.

## 3 L'accès lexical contextualisé

### 3.1 Les bases de connaissances

SVETLAN (De Chalendar et Grau, 2001) permet d'acquérir automatiquement des classes de noms par domaines sémantico-pragmatiques à partir de textes. Les domaines sont formés par l'agrégation de segments de textes similaires, segments obtenus par une analyse thématique (Ferret et Grau, 1998). Ils sont composés d'un ensemble de mots (noms simples et composés, verbes et adjectifs) possédant un poids qui représente leur degré d'importance pour le sujet décrit. Par application d'une analyse distributionnelle sur l'ensemble des segments de textes liés aux domaines, SVETLAN regroupe des mots jouant le même rôle syntaxique par rapport à un même verbe, où seuls les mots les plus pertinents pour décrire le domaine sont retenus. On obtient ainsi des domaines décrits par des ensembles de mots, mais aussi par des ensembles de structures « verbe, relation, classe de noms ». Une base de connaissance a été construite à partir d'un corpus d'articles de l'AFP d'environ 6000 dépêches découpées en 8000 segments thématiques et qui contient 1531 domaines. Pour le présent article, nous appellerons domaine l'ensemble des mots présents dans les structures d'un domaine.

Les informations produites par SVETLAN sont des données lexicales organisées conceptuellement, mais également indexées grammaticalement, puisque les noms sont liés aux verbes par des liens "Sujet", "COD" ou "Préposition" (à, sur, de, etc.). On a ainsi une approche phraséologique, une extension syntagmatique du terme à la phrase, grâce à la conservation des liens syntaxiques entre les noms et les verbes.

Les connaissances de SVETLAN ressemblent aux informations syntagmatiques et syntaxiques que possède un locuteur concernant un lemme. Ces ressources terminologiques sont toutefois limitées dans SVETLAN, mais elles sont extensibles de plusieurs façons, notamment par EWN.

EWN serait le pendant paradigmatique des informations concernant un lemme dans l'esprit d'un locuteur. Il comporte environ 32800 variants et pour chacun, au moins un lien paradigmatique vers un autre lemme.

## 3.2  Conception de l'expérimentation

Les différents types d'informations dont nous disposons représentent des données lexicales, stockées durablement en mémoire chez un locuteur. Cependant, lorsque l'on produit du langage, le choix d'un mot dépend du contexte phrastique.

Notre méthode consiste à utiliser des textes à trous (un article AFP avec 10 mots extraits) pour observer l'accès à un mot absent dans un (con)texte. L'ensemble du texte précédant et suivant le mot manquant est conservé sous forme lemmatisée (conservation des mots pleins uniquement). On obtient ainsi un environnement symbolisant la parole d'un locuteur dans un contexte, les mots manquants symbolisant le phénomène de TOT, les mots que l'on recherche. Afin d'évaluer la capacité de récupération des mots grâce à nos ressources, nous quantifierons nos résultats par les mesures du rappel et de la précision :

$$\frac{\sum (mot\,(s)\_trouvé\,(s))}{\sum (mot\,(s)\_recherché\,(s))} \qquad \frac{\sum (mot\,(s)\_trouvé\,(s))}{\sum (mot\,(s)\_renvoyé\,(s)\_total)}$$

Les ensembles de mots retrouvés par le système sont susceptibles d'être proposés au locuteur afin qu'il puisse y trouver le mot recherché.

## 3.3  Étude des liens non typés

On souhaite pouvoir identifier les informations pertinentes pour l'accès lexical au niveau des relations principalement. En effet, il s'agit d'observer dans quelle mesure il est important de typer les liens pour permettre l'accès à un mot. Les liens thématiques (syntagmatiques) et syntaxiques de SVETLAN suffisent-ils ? Une combinaison de liens syntagmatiques non typés avec des liens paradigmatiques typés comme ceux dont on dispose dans EWN amènent-ils un meilleur résultat ? Serait-ce profitable d'avoir un typage de tous ces liens (type Dictionnaire Explicatif et Combinatoire) ?

Notre application (cf. Fig.1) utilise les ressources mentionnées. Elle simule un locuteur (article AFP) se trouvant dans un état de TOT. Nous représentons ceci par les lemmes de ses mots pleins dans « Liste Contexte » et par le fait que plusieurs mots ont été ôtés du texte original (10 mots sont absents de son discours, voir Liste TOT). C'est en suivant les liens proposés par nos ressources (SVETLAN, EWN) que le locuteur est censé accéder au mot recherché (Liste TOT résolu).

En premier lieu, nous avons étudié l'accès simplement par liens associatifs à l'aide de la base SVETLAN (cf. fléchage 1, Fig.1). Cela consiste à mettre en correspondance les lemmes de la « Liste Contexte » avec les informations lexicales (Noms et Verbes) de SVETLAN. Cette mise en correspondance permet de sélectionner les domaines où au moins 75% des mots du texte sont présents. Il suffit ensuite de constater la présence ou l'absence de nos mots manquants dans les domaines sélectionnés (« Liste TOT résolu »).

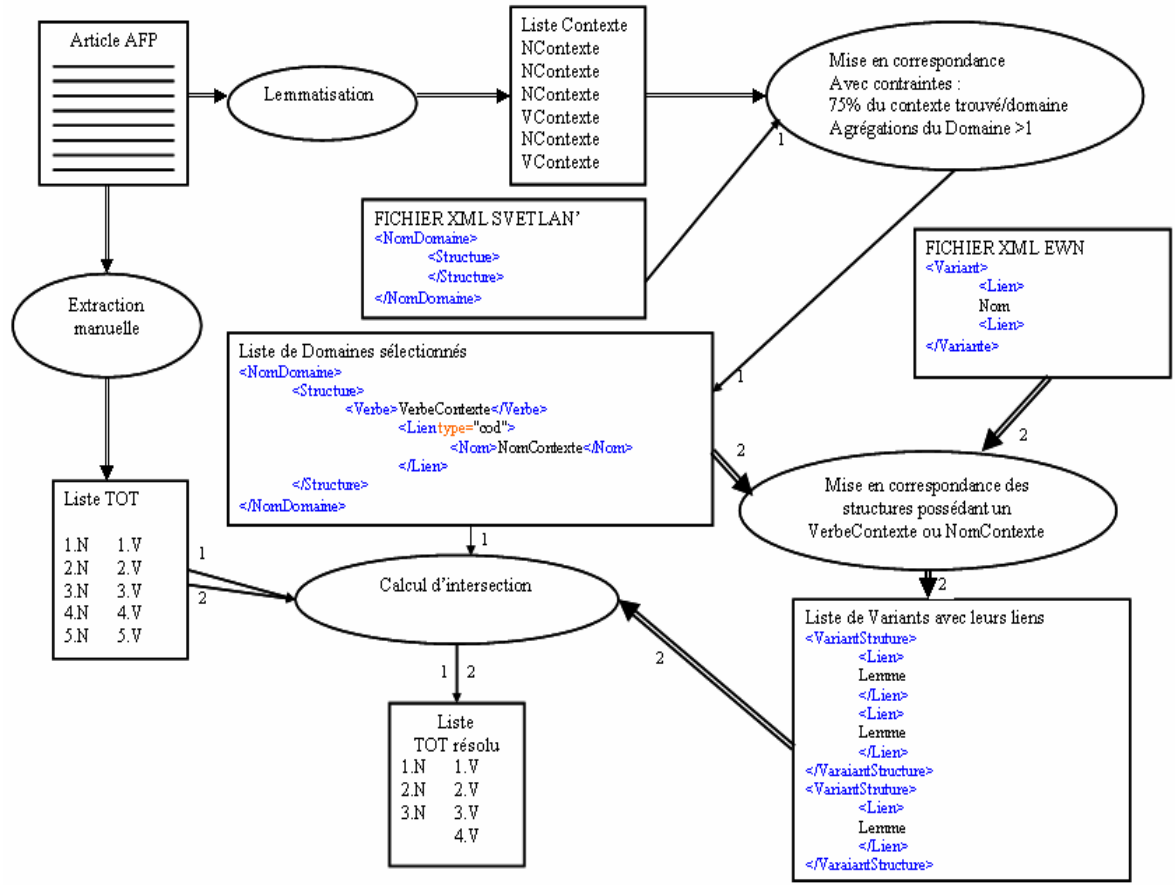

Figure 1. Méthode de résolution de TOT par liens syntagmatiques non typés et liens paradigmatiques typés à l'intérieur de structures

### 3.4 Apport des liens typés

La méthode précédente est raffinée, permettant en plus d'une recherche par des liens strictement associatifs, de prendre en compte les liens paradigmatiques des mots d'un domaine thématique en utilisant la base de données EWN (cf. fléchage 2, Fig.1). La démarche est identique à la précédente. Les mots des structures contenant au moins un mot du texte sont ensuite reliés aux variants d'EWN. On observe la présence des mots de la « Liste TOT » soit dans les domaines sélectionnés soit dans les mots reliés aux variants d'EWN correspondant à des mots d'un domaine sélectionné.

Il est possible d'effectuer une restriction contextuelle pour améliorer la précision, en découpant le texte en segments thématiques (STi dans Table 1). Moins de mots sont alors fournis au système, permettant de ce fait de baisser le bruit et d'améliorer la précision. Le rappel n'est pas touché par ces restrictions.

# 4   Résultats

## 4.1   Présentation des résultats

L'expérience décrite a été menée sur 4 textes de 225 mots en moyenne. Dans chacun de ces textes 10 mots ont été retirés. Nous avons choisi de retirer des noms et des verbes, car nos bases de connaissances contiennent essentiellement ces catégories lexicales. Outre des mots liés au contexte du texte (cf. Tab.1, militaire, partisan, enfreindre), des mots très généraux (cf. Tab.1, lundi, étudier, représenter) sont extraits.

Les tests effectués sur SVETLAN et EWN seuls offrent un rappel de respectivement 72,5% et 70 % en moyenne pour les 4 textes. La combinaison des deux ressources permet d'amener ce rappel à 92,5%. Les résultats présentés ici correspondent au détail de l'un d'eux (Irak), qui est représentatif de l'ensemble.

Selon le segment thématique du texte utilisé en tant que contexte, le nombre de domaines pris en compte pour le rappel des 10 mots est de 3 (pour un rappel de 7/10) à 7 (pour un rappel de 9/10).

Les domaines possédant 329 à 1709 mots pour 7 à 9 mots retrouvés, la précision est faible. Si l'on prend en compte la restriction par segment thématique, on constate que le segment thématique 3 apporte en général le rappel et la précision les plus importants. Cependant, le segment 1, représentant une introduction à l'article a une très bonne précision.

Au vu des résultats avec une restriction sur SVETLAN (cf. Tab.1), au niveau des structures, on constate que la précision est largement améliorée, puisque l'indice augmente en moyenne de 0.004 soit environ 266% sur l'ensemble des structures thématiques. En effet, on ne conserve que l'information pertinente proche.

## 4.2   Discussion des résultats

Cependant, malgré ces résultats encourageants, la précision est encore très faible puisque celle-ci représente la liste de mots dans laquelle le locuteur choisit le mot approprié. C'est à dire que celui-ci doit retrouver dans une liste d'environ 8000 mots (dans le cas d'Irak ou en moyenne 1600 mots par ST) liés à un nombre variable de mots dans EWN, les 10 qui lui manquent (alors que tous ne seront pas forcément disponibles). C'est alors que l'on peut combiner les différents types d'accès et sélectionner dans la liste des mots associés ceux de la catégorie lexicale cherchée, d'une position donnée dans la phrase (COD du verbe, sujet) ou encore des mots qui ressemblent phonétiquement à d'autres.

|  |  | ST1 | ST2 | ST3 | ST4 | ST5 |
|---|---|---|---|---|---|---|
| **Mots Trouvés** | 1. force | + | + | + | + | + |
|  | 2. lundi | + | + | + | + | + |
|  | 3. amende | - | + | + | + | + |
|  | 4. militaire | + | + | + | + | + |
|  | 5. partisan | - | + | + | - | + |
|  | 6. étudier | + | + | + | + | + |
|  | 7. représenter | + | + | + | + | + |
|  | 8. lancer | + | + | + | + | + |
|  | 9. confirmer | + | + | + | + | + |
|  | 10. enfreindre | - | - | - | - | - |

| | Nom du domaine | Nbre mots du domaine | Nbre mots avec structures seules + réduction du bruit | | Domaines sélectionnés par segment | | | | |
|---|---|---|---|---|---|---|---|---|---|
| **Domaines sélectionnés** | ApprendreDemander | 1015 | 386 | 62% |  | + |  | + | + |
|  | CommanderPoursuivre | 1709 | 1599 | 7,5% | + | + | + | + | + |
|  | DonnerObtenir | 956 | 212 | 78% |  |  |  | + | + |
|  | ÉlireEffectuer | 582 | 127 | 78% |  | + |  |  |  |
|  | FuirContrôler | 329 | 31 | 90,5% | + |  |  |  |  |
|  | MaintenirDemander | 810 | 117 | 85,5% |  |  |  |  | + |
|  | PrévoirAtteindre | 1017 | 112 | 89% |  |  |  |  | + |
|  | PrévoirReprendre | 1247 | 710 | 43% |  | + | + | + |  |
|  | TuerPrendre | 925 | 405 | 56% |  | + | + |  | + |
|  | TuerTrouver | 1309 | 862 | 34% | + | + | + | + | + |

Tableau 1. Résultats Occurrences de mots trouvés par SVETLAN+EWN dans STis(Irak)

Afin d'illustrer l'efficacité d'une amorce de type phonologique, nous avons interrogé notre base de connaissances avec les exemples suivants, sans restriction préalable par domaine. Si la phrase « l'Etat a ?*TOT ?* la loi » et que seul le mot « aboli » vient à l'esprit, il est possible de faire une mise en correspondance entre le contexte de la phrase, soit « loi » et les verbes proches de « abolir » phonologiquement (abaisser, abandonner, abasourdir, abâtardir, abattre, abcéder, abdiquer, abêtir, etc.). Le système trouve dans SVETLAN la cooccurrence « abroger une loi » et sera capable de proposer « abroger » pour résoudre le problème du TOT. Si l'on part d'un lapsus : « *mépriser* la situation », le système proposera alors « mettre » ou « maîtriser », mais il pourra également faire un choix par rapport à la fonction syntaxique du groupe « situation » (ici « lien = cod »). La base SVETLAN comportant : (1) « mettre » + « lien = dans » + « situation » et (2) « maîtriser » + « lien = cod » + « situation », on retiendra la seconde solution.

# 5 Conclusion

Cette première expérimentation montre que l'ensemble des relations construites automatiquement est utilisable pour accéder à un mot à partir de mots proches. La méthode utilisée pour construire en contexte des classes de noms associées syntaxiquement à des verbes permet de restreindre ces classes par domaine, mais aussi de les utiliser hors de restriction préalable par domaine. Les classes formées sont constituées de mots sémantiquement proches ayant des relations diverses telles que synonymie ou hyperonymie. La conservation de la relation syntaxique fournit un critère de sélection supplémentaire.

L'extension des classes par EWN s'est révélée productive montrant la compatibilité et la complémentarité des deux bases de connaissances. En effet, EWN ne disposant pas de relations syntagmatiques, il est souvent impuissant pour trouver le mot manquant. On notera que ces tests sont effectués avec des bases et des corpus de petites tailles ; on pourrait prévoir plusieurs types d'extension. Tout d'abord une extension de la base de domaines SVETLAN, où l'on fournirait d'autres corpus à traiter afin d'augmenter le nombre et la diversité des domaines disponibles. Parallèlement, on pourrait augmenter et préciser les liens paradigmatiques en utilisant des ontologies. Une extension supplémentaire pourrait être le typage des relations grâce au DEC. Cette ressource pourrait faire partie d'un système plus important ou tout simplement être l'extension d'un dictionnaire électronique existant (TLF, WordNet).

Quoi qu'il en soit, il semblerait qu'une ressource comme SVETLAN pourrait être un élément nécessaire pour construire une application sémantique basée sur l'association et destinée à aider un être humain à trouver le mot qu'il a sur le bout de la langue.

# 6 Références